\documentclass[letterpaper]{article} 
\usepackage{aaai25}  
\usepackage{times}  
\usepackage{helvet}  
\usepackage{courier}  
\usepackage[hyphens]{url}  
\usepackage{graphicx} 
\usepackage{natbib}  
\usepackage{caption} 
\usepackage{algorithm}
\usepackage{algorithmic}

\usepackage{newfloat}
\usepackage{listings}
\DeclareCaptionStyle{ruled}{labelfont=normalfont,labelsep=colon,strut=off} 
\floatstyle{ruled}
\newfloat{listing}{tb}{lst}{}
\floatname{listing}{Listing}
\usepackage{amsmath}
\usepackage{amssymb}
\usepackage{multirow}
\usepackage{pifont}

\title{VOILA: Complexity-Aware Universal Segmentation of CT images \\ by Voxel Interacting with Language}
\author {
    Zishuo Wan\textsuperscript{\rm 1},
    Yu Gao\textsuperscript{\rm 1},
    Wanyuan Pang\textsuperscript{\rm 1},
    Dawei Ding\textsuperscript{\rm 1, \rm 2}\thanks{Corresponding author}
}
\affiliations {
    \textsuperscript{\rm 1}School of Automation and Electrical Engineering, University of Science and Technology Beijing, Beijing 100083, China\\
    \textsuperscript{\rm 2}Key Laboratory of Knowledge Automation for Industrial Processes, Ministry of Education, Beijing 100083, China\\
    d202410372@xs.ustb.edu.cn, dingdawei@ustb.edu.cn
}

\begin{document}

\maketitle

\begin{abstract}
Satisfactory progress has been achieved recently in universal segmentation of CT images. Following the success of vision-language methods, there is a growing trend towards utilizing text prompts and contrastive learning to develop universal segmentation models. However, there exists a significant imbalance in information density between 3D images and text prompts. Moreover, the standard fully connected layer segmentation approach faces significant challenges in handling multiple classes and exhibits poor generalizability. To address these challenges, we propose the VOxel Interacting with LAnguage method (VOILA) for universal CT image segmentation. Initially, we align voxels and language into a shared representation space and classify voxels on the basis of cosine similarity. Subsequently, we develop the Voxel-Language Interaction framework to mitigate the impact of class imbalance caused by foreground-background discrepancies and variations in target volumes. Furthermore, a Complexity-Aware Sampling method is proposed to focus on region hard to segment, achieved by generating pseudo-heatmaps from a trainable Gaussian mixture distribution. Our results indicate the proposed VOILA is capable to achieve improved performance with reduced parameters and computational cost during training. Furthermore, it demonstrates significant generalizability across diverse datasets without additional fine-tuning.
\end{abstract}

%
\begin{links}
    \link{Code}{https://github.com/ZishuoWan/VOILA}
\end{links}

\section{Introduction}
\label{sec:introduction}
The accurate segmentation of anatomical structures in medical images is a fundamental task in clinical practice and biomedical research, including diagnosis, treatment planning, and the monitoring of disease progression. However, manual segmentation is labor-intensive, time-consuming, and in need of expertise, necessitating the development of automated segmentation methods. With the success of UNet \cite{ronneberger2015u} and its variants \cite{chen2021transunettransformersmakestrong, cao2022swin}, the end-to-end deep learning models has become the baseline for the segmentation. 

However, most existing models lack generalization, requiring separate training on each dataset to achieve good performance on their respective test sets. This approach is significantly less meaningful, compared to training a single model on one dataset that can perform well across multiple datasets. The issue is further compounded in methods that necessitate separate training sessions for each organ, which is even less efficient and practical. Since the introduction of Contrastive Language-Image Pre-training \cite{radford2021learning}, which marked a milestone in combining the modalities of computer vision and natural language, there has been substantial success across several vision-language tasks \cite{li2022languagedriven, gu2022openvocabulary, ramesh2022hierarchicaltextconditionalimagegeneration}. By constructing a text prompt using a template following a certain pattern, such as \emph{"A photo of a \{label\}"} where \emph{\{label\}} is typically filled with the class name, every image input is paired with a corresponding language input. This approach allows traditional visual tasks to benefit from the additional input dimension provided by language. However, the essence of contrastive learning lies in the one-to-one pairing between vision and language inputs. Consequently, text prompts constructed from templates cannot achieve the uniqueness required, limiting their effectiveness in visual-only tasks. Moreover, the information density of a template is significantly lower than that of an image, further hindering its broader application.

In medical segmentation tasks, the objective is to establish a mapping function, where pixels (or voxels, in the case of 3D CT images) are assigned to specific categories. This process typically involves two steps: (\romannumeral1) Extracting a hidden representation for each voxel; (\romannumeral2) Classifying the voxel based on its hidden representation. In encoder-decoder structured deep learning models, these steps are executed consecutively, where the encoder extracts a hierarchical high-level representation and the decoder maps it to the class probabilities of each voxel. However, most models place a strong emphasis on the first step by increasing the complexity of the encoding process or replacing the encoder with new architectures such as TransUNet\cite{chen2021transunettransformersmakestrong}, Swin-UNet\cite{cao2022swin} and Swin-UNetr\cite{hatamizadeh2021swin}, while the second step is often simplified to a basic fully connected layer. In models like the UNet series, where skip connections and hierarchical decoders are employed, these components primarily serve to further refine the encoded representation. The classification layer, which is fully data-driven and composed solely of learnable parameters, sees its computational burden scale with the image size, the dimension of the representation, and the number of classes. In universal segmentation models, as the number of classes increases, the computational cost of this layer grows linearly. Moreover, the inclusion of a large number of unnecessary background voxels in the computation not only leads to significant computational inefficiency but also causes foreground voxels to be overshadowed by the background during the training stage.

To address these challenges, we propose VOxel Interacting with LAnguage method (VOILA), a brand new approach for multi-organ segmentation. We designed a voxel-text representation framework from a voxel-centered perspective. By employing cosine similarity, voxels and text tokens are mapped into the same feature space, with similar categories drawn closer and dissimilar ones pushed farther apart. Several strategies are employed to mitigate class imbalance caused by foreground-background discrepancies and variations in target volumes, while also enhancing the generalizability of the model. Additionally, we introduce a Complexity-Aware Sampling (CAS) module that leverages self-supervised learning during training. It dynamically selects regions with higher segmentation difficulty for reinforcement, thereby accelerating model convergence and achieving strong performance with fewer parameters and lower computational costs. The main contributions of this work are summarised as follows:
\begin{enumerate}
    \item To the best of our knowledge, we are the first to introduce voxel-wise contrastive learning into segmentation. 
    \item We develop a Voxel-Language Interaction framework VOILA based on cosine similarity for generalizable universal segmentation.
    \item We propose a self-supervised Complexity-Aware Sampling module that models voxel-level complexity using a Gaussian mixture distribution and intensively trains the model on hard-to-segment regions.
    \item The proposed method achieves competitive performance on 7 public datasets with lower computational cost and demonstrates remarkable generalization ability.
\end{enumerate}

\section{Related Work}
\label{sec:related_work}
\subsection{Vision-Language Segmentation}
Many studies have further validated the significant performance of contrastive learning techniques in language-image pre-training \cite{mu2022slip,singh2022flava,yu2022coca}. CLIP utilizes an image-text dual-stream encoder to learn joint visual-language representations by projecting encoded images and text into a shared embedding space, demonstrating substantial potential for image segmentation applications \cite{shin2022reco,wang2022cris,zhou2023zegclip}. Subsequent works have expanded on the CLIP framework. For example, DenseCLIP \cite{rao2022denseclip} and LSeg \cite{li2022languagedriven} extend this paradigm to dense prediction tasks, achieving outstanding results in semantic segmentation. RegionCLIP \cite{yi2023simple} enhances CLIP’s image input to learn region-level visual representations. SimSeg \cite{zhong2022regionclip} employs locality-driven alignment (LoDA) strategies to address non-contextual information alignment issues. Additionally, efficient image segmentation can be achieved through methods such as inter-modal cross-attention \cite{lee2023text}, joint feature learning with masked image/language modeling, and cross-modal alignment losses \cite{chng2024mask}. In this paper, we adopt a voxel-centric approach, exploring how interactions between voxel tokens and text tokens can determine the category of each voxel.

\subsection{Universal Segmentation Models}
To achieve high generalization performance, the most straightforward approach is to develop larger and more diverse datasets \cite{ulrich2023multitalent,moor2023foundation} or to create scalable and transferable deep learning models \cite{huang2023stu} for pre-training, aiming to maintain strong segmentation performance on unseen datasets. These datasets can include various medical modalities, such as medical imaging, electronic health records, laboratory results, genomics, graphs, or medical texts \cite{moor2023foundation}. Multimodal data provides prior knowledge, often detailing anatomical structures or imaging patterns before further image processing. To capture anatomical relationships effectively, several strategies can be employed in segmentation models. For instance, DoDNet \cite{zhang2021dodnet} incorporates task indices as one-hot vectors for additional model input. Other studies integrate different modalities as prompts within the feature space \cite{ye2023uniseg,butoi2023universeg,qin2023medical} or fine-tune SAM models for universal medical image segmentation \cite{zhang2023customized,gao2023desam}. Additionally, structured text features, when combined with CLIP-driven methods \cite{liu2023clip,wang-etal-2022-medclip}, can be embedded into segmentation models. Incremental learning has also been explored for its advantages in universal medical models \cite{yi2023towards}. While these methods inevitably require a large number of parameters and computational resources to train the model, this work aims to achieve competitive performance with significantly lower cost.

\section{Method}
\label{sec:method}
\subsection{Overall Architecture}

\begin{figure*}[t!]
\centering
\includegraphics[width=0.9\textwidth]{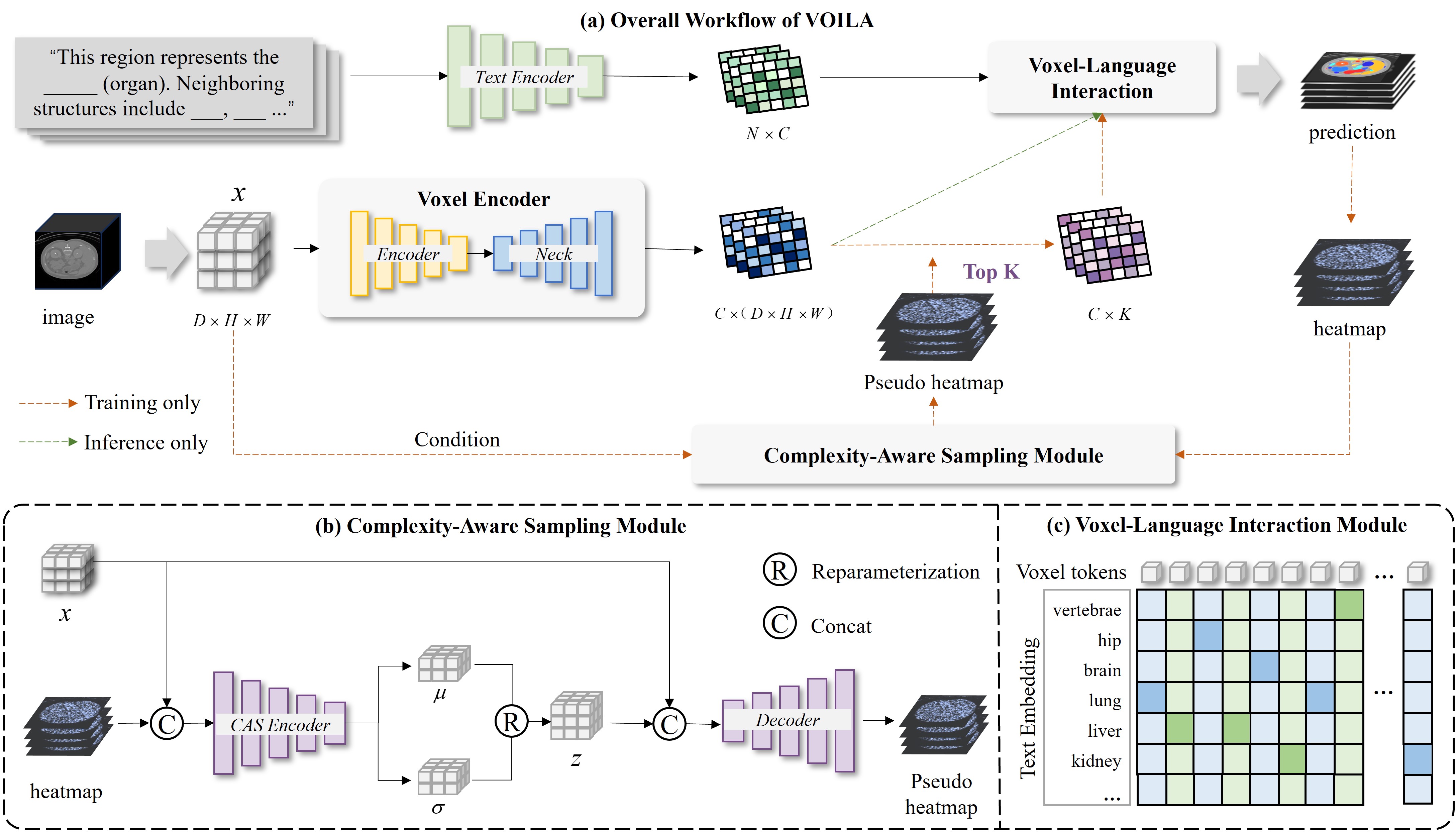} 
\caption{Overview of VOILA. (a) The overall workflow of VOILA. When taking CT images and text prompts as inputs, the encoders extract their representation tokens respectively. The voxels are selected by (b) Complexity-Aware Sampling module. Finally, the tokens interact across modalities in (c) Voxel-Language Interaction module for classification.}
\label{overview}
\end{figure*}

The overall flow of VOILA is shown in Figure \ref{overview}. The entire architecture consists four components: voxel encoder, text encoder, complexity-aware self-supervised sampling module, and voxel-language interacting module. The voxel encoder is employed to obtain the representation of each voxel. This representation is not only determined by its own grayscale value but should also incorporate information from neighboring voxels to accurately describe its relative position and context within the image. Convolution is more effective than self-attention in capturing local information, as it avoid diluting the specificity of a voxel with excessive global information. Furthermore, during the tokenization process in ViT, aggregating information within a patch to reduce computational load is counterproductive for detailed voxel characterization. In this regard, we use residual convolution modules instead of self-attention modules to construct the hierarchical backbone network for feature extraction. Given that features are multiscale, we include a Feature Pyramid Network (FPN) neck after the backbone to fuse multiscale representation tokens, which forms the complete voxel encoder together with the backbone. When a CT image is input, the voxel encoder produces a \emph{hash table}, where the key represents the position of each voxel in the image, and the value is the c-dim token for the particular voxel. This hash table is then used in the sampling  process, which will be detailed in the subsequent section. The text encoder leverages the pre-trained CLIP model, enabling efficient extraction of features from text prompts for classification, similar to the zero-shot inference process in CLIP.

\subsection{Voxel-Language Interaction: A Voxel-Centric Perspective}
First, we constructed text templates for each class, typically formatted as \emph{"This region represents the \{label\}"}. The text encoder generates text representation tokens for these prompts. Unlike existing approaches that use text features as image-wise auxiliary decision aids, this paper employs them as the basis for voxel-wise decision-making. Once the hash table is obtained, the corresponding representation token for each voxel can be retrieved using its coordinates. For each voxel, we compute the cosine similarity between its token and each text token. The class associated with the text token that has the highest similarity is considered the classification result for that voxel. Specifically, when calculating cosine similarity, supposing an image with $D \times H \times W$ voxels and $N$ segmentation classes, there will be $D \times H \times W$ positive samples and $D \times H \times W \times (N-1)$ negative samples, assuming no sampling. This setup can be optimized using cross-entropy loss:
\begin{equation}
\mathcal{L}_{v \to \{ t_i \} }=-\log \frac{\exp \left(v \cdot t_+ / \tau\right)}{\sum_{i=1}^N \exp \left(v \cdot t_i / \tau\right)}
\end{equation}
where $v \in  \mathbb{R}^{1 \times C} $ and $t \in  \mathbb{R}^{1 \times C} $ are $C$-dim representation tokens for voxels and texts respectively, and $\tau$ is a temperature hyper-parameter like InfoNCE \cite{oord2019representationlearningcontrastivepredictive}. 

\begin{figure}[t!]
\centering
\includegraphics[width=\columnwidth]{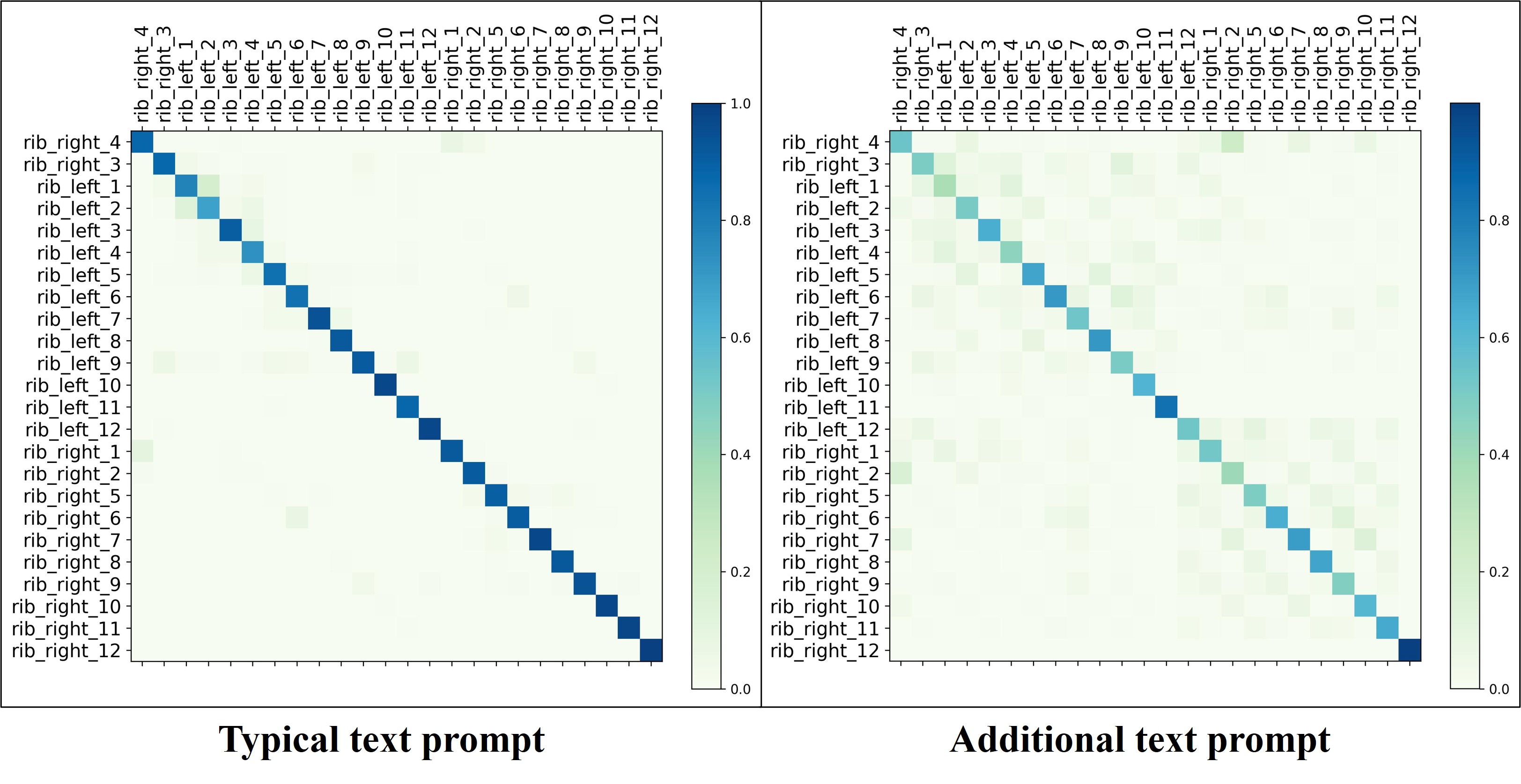} 
\caption{Cosine similarities of text tokens extracted for the text encoder. The additional text prompt in this paper include more cross-text interactions. }
\label{textsimilarity}
\end{figure}

\begin{figure}[t!]
\centering
\includegraphics[width=\columnwidth]{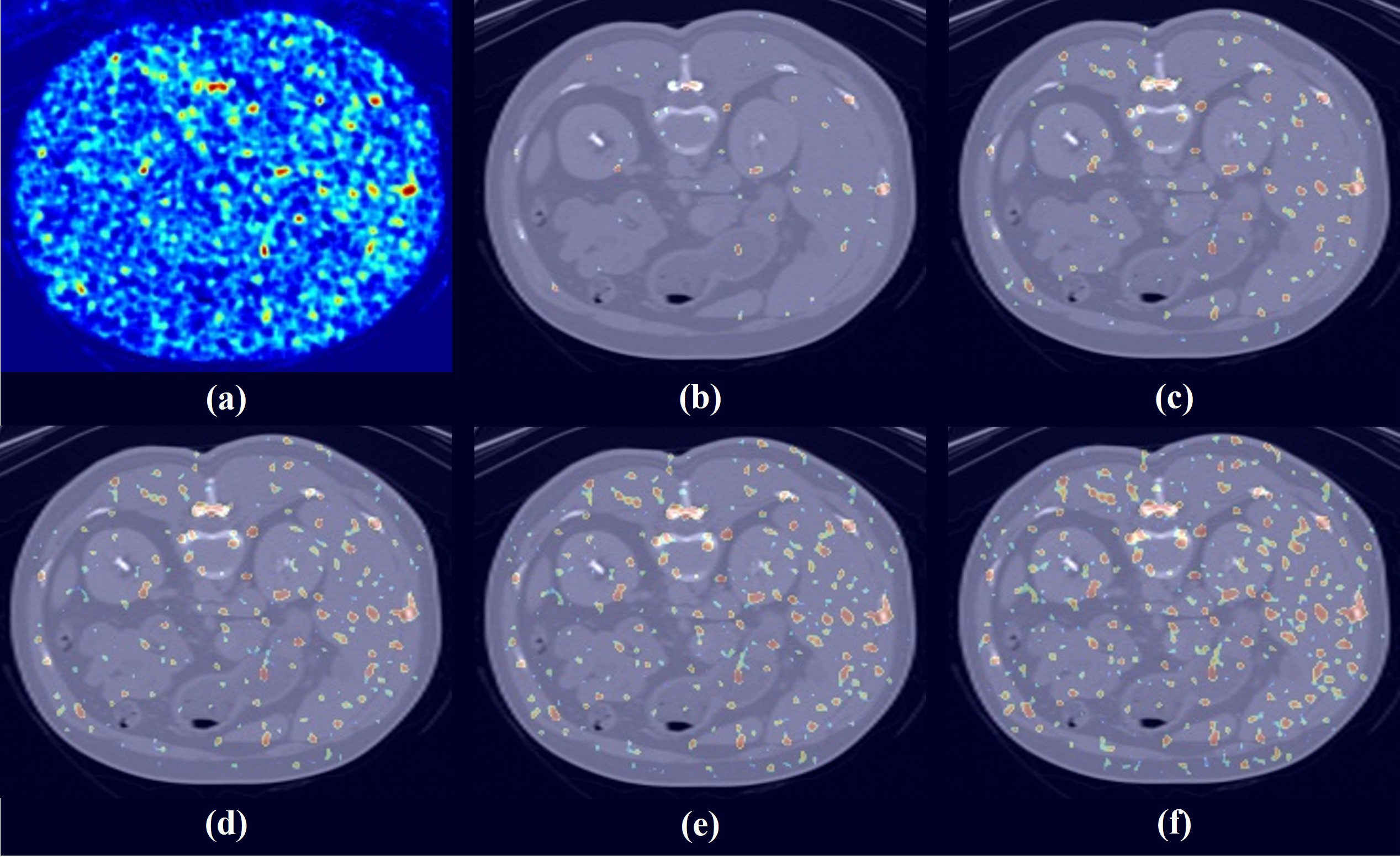} 
\caption{(a) The pseudo heatmap generated by the CVAE in the Complexity-Aware Sampling module. Then the CAS module samples voxels with different sampling rate (b)-(f).}
\label{sampling}
\end{figure}

\subsubsection{Cross-Text Interaction}
Unlike the CLIP training process, where images and texts have a one-to-one correspondence, template approach only ensures that each voxel has a unique corresponding text but not vice versa. As a result, cosine similarity calculations involve only voxel-to-text interactions, with no text-to-voxel interactions under the circumstances. So the cross-entropy is asymmetric and there is only one direction $v \to \{ t_i \}$. In addition, the original templates differed only by the class term, which limits the variability of the extracted features to a single word and reduces separability in the feature space. To introduce some interaction between voxels of different categories and enhance separability, we improved the text prompt template. We incorporated spatially neighboring organs or structures with prompts like \emph{"Neighboring structures include \{STR1\},  \{STR2\} ..."}, to encourage spatially related structures to be closer in the feature space, while unrelated structures are positioned farther apart. The cosine similarities in the text representation tokens are partially visualized in Figure \ref{textsimilarity}.

\subsubsection{Computational Complexity}
Classifying $DHW$ voxels into $N$ categories can be viewed as matrix multiplication. Assuming the dimension of both voxel tokens and text tokens is $C$, the total computational complexity for calculating cosine similarity is 
\begin{equation}
   \Omega(C) = DHWCN,
\end{equation}
which is equivalent to the computation cost for classifying voxels with a fully connected layer. However, in our case, we can first significantly reduce the dimension $C$ to $M$. As a result, when performing the matrix multiplication, the computational complexity becomes
\begin{equation}
\Omega(M) = DHWCM + NCM + DHWMN.   
\end{equation}
Since $DHW$ is significantly large and $M \ll C$, it turns out to be more affordable. 

\subsubsection{Class Imbalance Problem}
The presence of numerous background points and differences in target volumes can overshadow smaller targets. Since we decompose the image into discrete voxels, the original Dice loss, which relies on intersection and union, is no longer applicable. Therefore, we adapted Dice loss into a voxel-wise F1 loss to mitigate the inherent class imbalance issue:
\begin{equation}
\begin{aligned}
\mathcal{L}_{\mathrm{F1}} &= 1-\frac{1}{N}\frac{\rm 2\cdot TP}{\rm 2\cdot TP+FP+FN} \\
&= 1-\frac{1}{N} \frac{2\cdot o_+}{1+o_+ + \sum_{i \in N} o_i \cdot (1-g_i)}
\end{aligned}
\end{equation}
where $g$ is an one-hot label vector for each voxel and $N$ means only the foreground classes. 

\begin{table*}[t!]
\centering
\begin{tabular}{c|c|ccccccc}
\multirow{2}{*}{Method} & Trainable & Ts-v2         & WORD          & AMOS & BTCV & Ab-1K & LiTS & Pancreas \\
                        & Params(M) & (117)         & (16)          & (15) & (13) & (4)   & (1)  & (1)      \\ \hline
nnUNet \cite{isensee2021nnu}                  & 22.68     & \underline{87.9}    & \underline{81.1}    & \textbf{84.0} &   70.9   & \underline{92.6}  & 90.9 & 75.1     \\
UNETR++ \cite{shaker2024unetr++}                & 42.97     & 87.0          & 76.5          & 82.5 & \underline{73.4} & \textbf{93.1}  & \textbf{94.8} & \textbf{75.5}     \\
nnFormer \cite{zhou2023nnformer}               & 149.46    & 64.4          & 80.1          &  80.0    & 68.0 &   89.5    & 91.7 & \underline{75.3}     \\ \hline
VOILA                   & \textbf{6.44}      & \textbf{92.1} & \textbf{83.0} & \underline{83.4} & \textbf{74.1} & 92.0  & \underline{91.9} & 73.3    \\
\end{tabular}
\caption{Comparison with 3 SOTA methods on 7 public datasets after 400 training epochs. The results are evaluated with average Dice score. The values in the second column only account for the number of parameters optimized during training, excluding frozen parameters. The numbers in brackets below the dataset names indicate the number of foreground classes.}
\label{table1}
\end{table*}

\subsection{Complexity-Aware Self-Supervised Sampling}
Since voxels are treated as independent entities for calculating cosine similarity and cross-entropy, it is possible to filter them fitting a certain pattern. It is obvious that optimizing with a large number of background points is detrimental to training efficiency. Additionally, voxels at the borders of an organ are generally more challenging to classify correctly compared to those in the interior. To address these issues, we propose a self-supervised sampling method that reduces computational costs while focusing on the most informative voxels during training. 

We assume that the classification complexity of each voxel can be quantified by a mixture of $g$ univariate Gaussian distributions, and these values can form a heatmap-like image. Therefore, we construct a lightweight conditional variational auto-encoder (CVAE) to fit this distribution. When taking a CT image as conditional input, the auto-encoder generates a corresponding pseudo heatmap $H \in \mathbb{R}^{D \times H \times W} $ to assess complexity and guide the sampling process. The self-supervised sampling process of the Complexity-Aware Sampling (CAS) module is detailed below in terms of self-supervised training and sampling.

\subsubsection{Self-Supervised Training}
Assuming that all voxel tokens in the image have interacted with the text tokens, we can obtain a classification confidence for each voxel. By sorting the voxels based on this confidence, we derive a complexity order, which also reflects uncertainty. If we assign values to the voxels in descending order from 1 to 0 and then smooth this map with a Gaussian filter, we obtain a heatmap that reflects the complexity. The complexity heatmap can be used to train the CVAE with reconstruction loss and KL divergence loss. The CVAE uses the heatmap as the reconstruction target and the CT image as the conditional input. First, both are jointly mapped into another hash table, where the keys are also the voxel coordinates, and the values correspond to the mean $\mu \in \mathbb{R}^{1 \times g}$ and variance $\sigma \in \mathbb{R}^{1 \times g}$ vectors at those positions. Next, with the reparameterization trick, the $g$ variables sampled from the standard Gaussian distribution are transformed into $g$ variables sampled from $g$ different Gaussian distributions. These Gaussian variables are then added with CVAE decoder. Finally, a sigmoid function is applied to constrain the output values between 0 and 1 to the reconstruct the heatmap.

\begin{table*}[t!]
\centering
\begin{tabular}{c|ccc|ccc}
 Dataset & \multicolumn{3}{c|}{w/o Fine-tuning} & \multicolumn{3}{c}{Supervised} \\
         & Dice    & NSD     & HD95   & Dice     & NSD      & HD95     \\ \hline
BTCV     & \textbf{81.4(+7.3)}    & \textbf{80.9(+9)}    & \textbf{16.0(-22.2)}   & 74.1     & 71.9     & 38.2     \\
Pancreas & \textbf{82.4(+9.1)}    & \textbf{80.2(+10.2)}    & \textbf{10.5(-12.2)}   & 73.3     & 70.0     & 22.7     \\
WORD     & 81.2(-1.8)    & 71.7(-7.5)    & \textbf{20.6(-4.7)}   & \textbf{83.0}     & \textbf{79.2}     & 25.3     \\
LiTS     & 91.8(-0.1)    & \textbf{80.9(+4.5)}    & \textbf{39.3(-10.5)}   & \textbf{91.9}     & 76.4     & 49.8     \\
Ab-1K    & 90.5(-1.5)    & \textbf{80.2(+6.9)}    & \textbf{16.6(-13.7)}   & \textbf{92.0}     & 73.3     & 30.3     \\
AMOS     & 77.1(-6.3)    & 72.4(-3.7)    & \textbf{16.7(-3.4)}   & \textbf{83.4}     & \textbf{76.1}     & 20.1     \\
\end{tabular}
\caption{Comparison of results on 6 datasets. Left: VOILA trained on the Ts-v2 dataset and inferred on testsets without fine-tuning. Right: VOILA trained and inferred separately on each dataset.}
\label{table2}
\end{table*}

\begin{figure}[t!]
\centering
\includegraphics[width=\columnwidth]{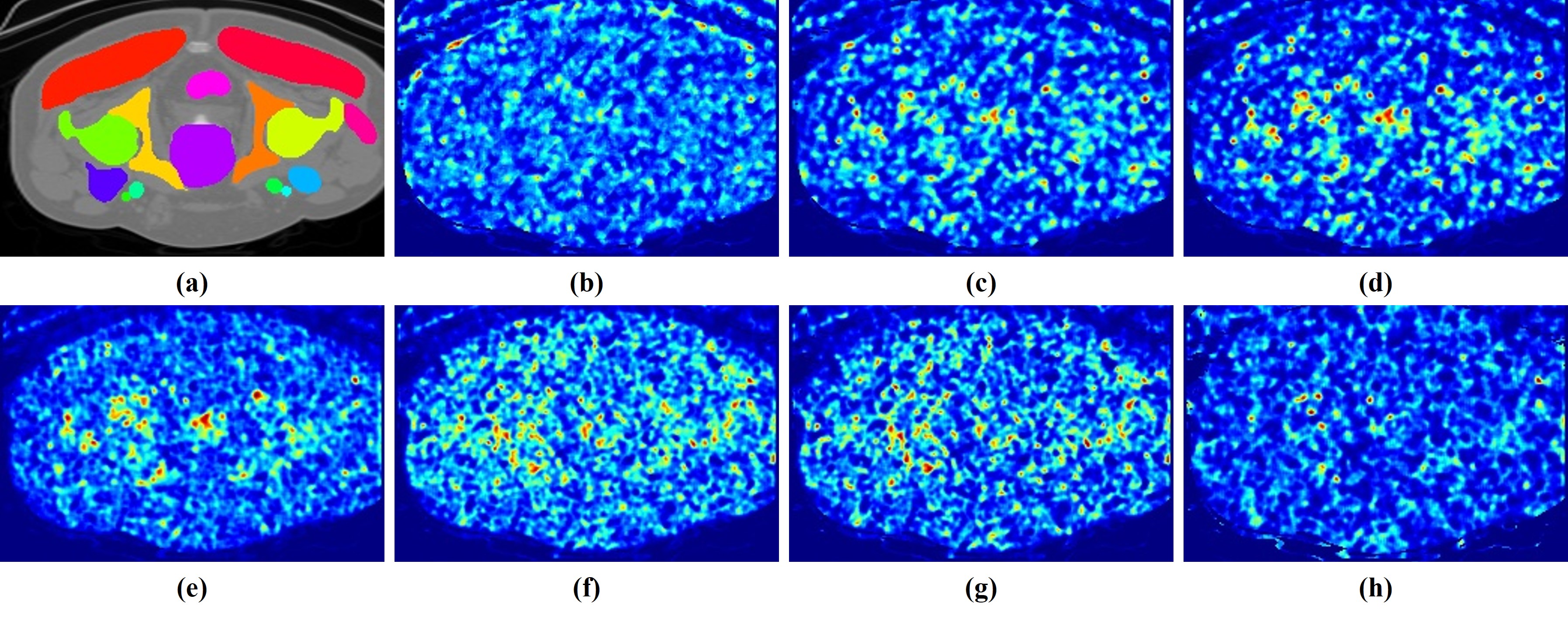} 
\caption{Example results for heatmap generated by the CVAE in the CAS module. (a) The groundtruth label. Heatmaps (b)-(f) are selected sequentially throughout the entire training process. The entire training phase involves a sampling process that begins with a randomly discrete pattern, gradually aggregates at key locations, and then disperses into finer details.}
\label{heatmaps}
\end{figure}

\begin{figure*}[t!]
\centering
\includegraphics[width=0.7\textwidth]{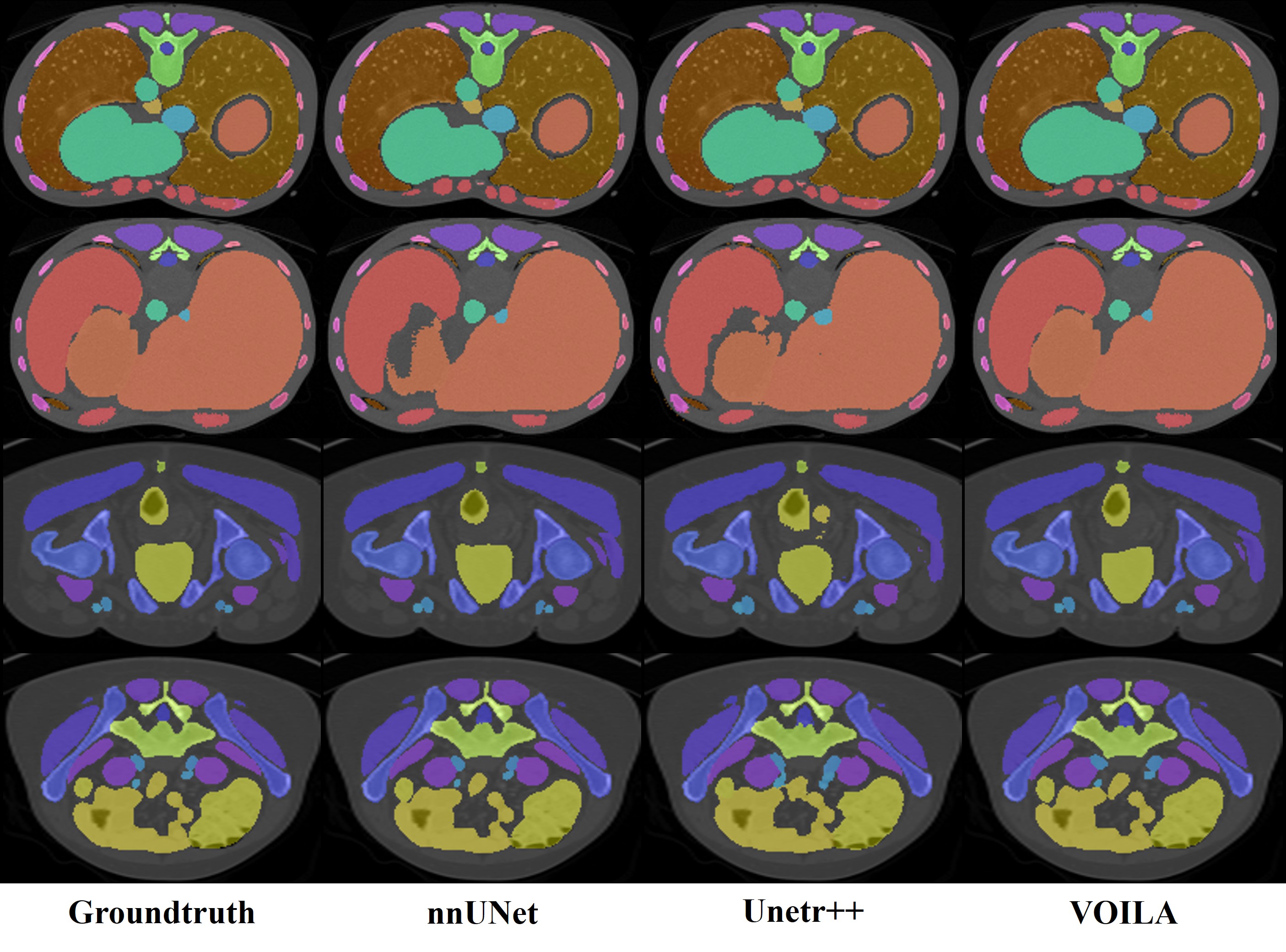} 
\caption{The visual comparison of 3 methods on Totoalsegmentator-v2.}
\label{results}
\end{figure*}

\begin{table*}[t!]
\centering
\begin{tabular}{c|cc|cc|ccc|ccc}
\multirow{2}{*}{V-L Interaction} & \multicolumn{2}{c|}{Sampling} & \multicolumn{2}{c|}{Ts-v2}    & \multicolumn{3}{c|}{BTCV}                     & \multicolumn{3}{c}{WORD}                      \\
                                 & Method         & Ratio        & Dice          & HD95          & Dice          & NSD           & HD95          & Dice          & NSD           & HD95          \\ \hline
\ding{52}                              & CAS            & 0.1          & \textbf{92.1} & \textbf{11.2} & \textbf{81.4} & \textbf{80.9} & \textbf{16.0} & \textbf{81.2} & \textbf{71.7} & 20.6          \\
\ding{52}                              & CAS            & 0.01         & 88.6          & 18.0          & 77.7          & 76.7          & 26.8          & 78.5          & 68.9          & 23.6          \\
\ding{52}                              & \ding{56}             &              & 86.0          & 20.5          & 73.0          & 67.7          & 46.4          & 73.5          & 60.1          & 50.4          \\
\ding{52}                              & Random         & 0.1          & 91.5          & 12.8          & 79.9          & 79.5          & 19.5          & 80.4          & 71.5          & \textbf{19.5} \\
\ding{56}                               & CAS            & 0.1          & 88.1          & 23.8          & 77.3          & 75.6          & 43.7          & 77.6          & 67.4          & 32.6                \\
\end{tabular}
\caption{Ablation results of proposed methods. All models were trained on Ts-v2 dataset and inferred without fine-tuning on BTCV and WORD to show the generalizability.}
\label{table3}
\end{table*}

\subsubsection{Complexity-Aware Sampling}
During the sampling process, random noise $Z \in \mathbb{R}^{g \times D \times H \times W} $ is drawn from standard Gaussian distribution and fed into the CVAE decoder along with the conditional CT image to generate a corresponding pseudo heatmap. This pseudo heatmap is used to represent classification complexity. The $K$ voxels with the highest complexity are then selected based on their coordinates. Finally, the sampling process is completed by retrieving the corresponding voxel tokens from the representation hash table and their labels from the ground truth, which are involved in Voxel-Language Interaction and calculating the cosine similarity. Once the complexity is obtained, a heatmap can be generated to guide the self-supervised reconstruction of the CVAE. By sampling, the computational complexity is reduced to 
\begin{equation}
    \Omega(M, K) = KCM + NCM + KMN,
\end{equation}
which significantly lowers the cost further compared with (3). 

\subsubsection{Avoiding Self-Loop}
However, the true complexity can only be determined after calculating the cosine similarity. If only the cosine similarity of these sampled points is calculated and the point with the lowest confidence is selected, the sampling module may fall into a self-reinforcing loop. In this scenario, the ranking of sampled points can be seen as a local optimal ground truth. Using this local optimal as the target for optimizing the CAS module results in a process of seeking local optima within local optima, causing the sampling outcome to deviate further from the global optimum. Therefore, during the self-supervised training, we randomly oversampled $nK$ voxels from a uniform distribution. These voxels, along with those sampled by the CAS module, interact with the text to construct a heatmap, which guides the optimization of the CAS. When $n=0 $, the process is equivalent to sampling only by CAS, while when $ n $ is sufficiently large, it becomes equivalent to completely random sampling.

\subsection{Optimization Objective}
In summary, the optimization objective of this paper consists of two parts: voxel-text interaction and self-supervised sampling. Therefore, the total loss function is written as:
\begin{equation}
    \mathcal{L} =  \sum_{v=1}^K (\mathcal{L}_{v \to \{ t_i \} } +  \mathcal{L}_{\mathrm{F1}}) + \mathcal{L}_{\mathrm{MSE}}(H,H') + \lambda \mathcal{L}_{\mathrm{KLD}}
\end{equation}
where $\lambda$ is the hyper-parameter, $H$ and $H'$ are original and reconstructed heatmap respectively.

\section{Experiments and Results}
\subsection{Datasets}
We conduct the experiments on 7 public CT datasets:
\begin{itemize}
    \item Totalsegmentator v2 \cite{wasserthal2023totalsegmentator} (Ts-v2)
    \item BTCV \cite{btcv2015}
    \item Pancreas-CT \cite{roth2016pancreas}
    \item WORD \cite{luo2022word}
    \item LiTS \cite{bilic2023liver}
    \item AbdomenCT-1K \cite{Ma-2021-AbdomenCT-1K} (Ab-1K)
    \item AMOS \cite{ji2022amos}
\end{itemize}
All datasets were randomly split into training and testing sets with a 1:1 ratio, except for the Totalsegmentator dataset, which used the official split. Dataset details can be found in Appendix.

\subsection{Implementation Details}
All experiments were conducted using the PyTorch platform and trained/tested on 8 NVIDIA GeForce RTX 3090 GPUs. All images were pre-processed by: resampling to a spacing of (1.5, 1.5, 1.5), crop to non-zero area and Z-score normalization. The networks were trained 400 epochs using the Adam optimizer with a learning rate of $1 \times 10^{-3}$. Parameters from pre-trained CLIP text encoder were frozen. Data augmentation was applied including randomly flipping, rotating, zooming, intensity adjusting, and patch crop with size of $128\times 128 \times 128$. As a result, sliding window prediction was performed during inference. The dimension of voxel and text tokens were first reduced to 32 before interaction. In terms of CAS module, we sampled 10\% voxels during training, with the oversampling ratio $n=2$. Following existing works, the Dice score, Normalized Surface Dice (NSD) and Hausdorff Distance (HD95) were utilized for quantitative comparison.

\subsection{Comparison with the State-of-the-Arts}
\begin{figure}[t!]
\centering
\includegraphics[width=0.75\columnwidth]{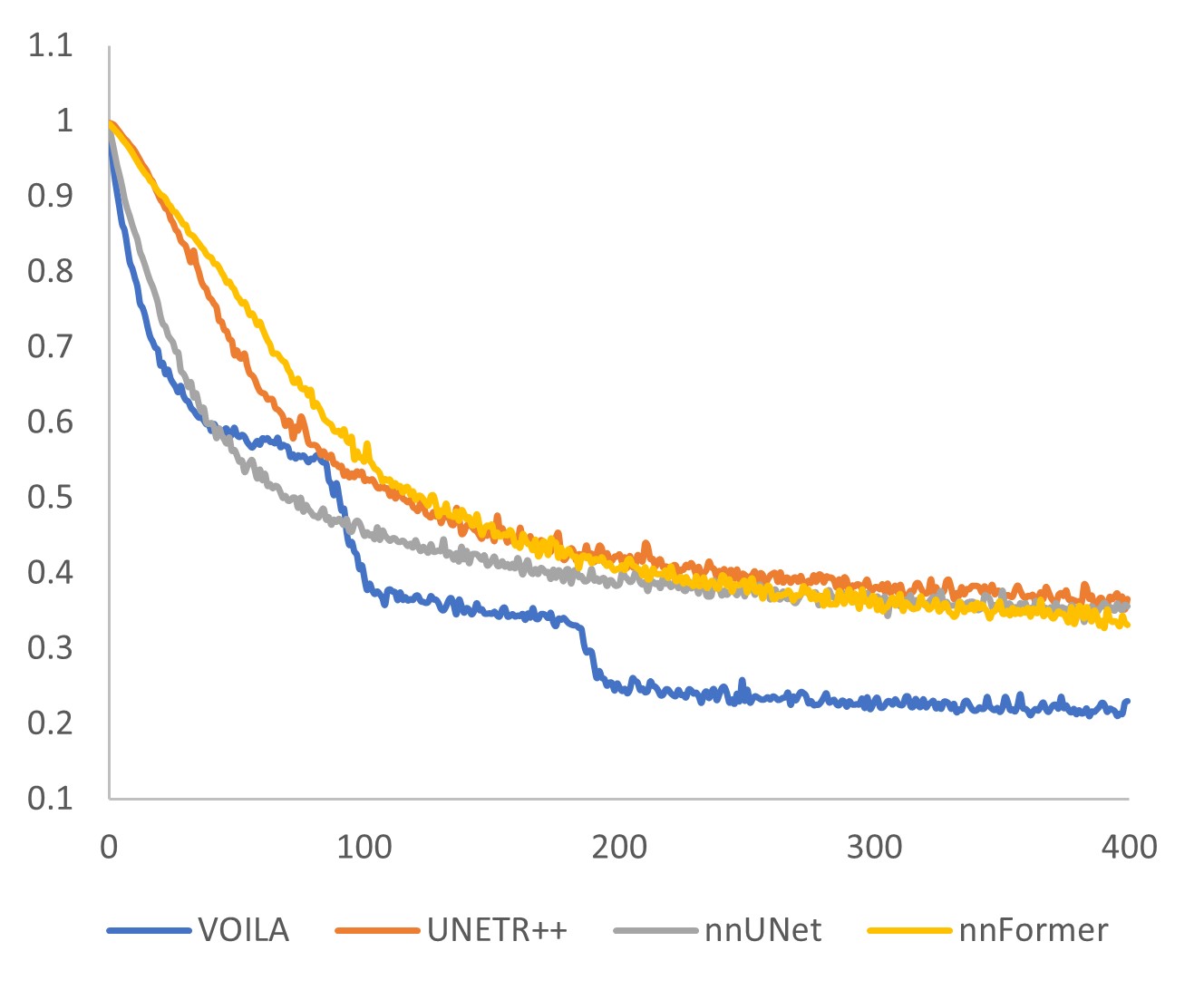} 
\caption{The voxel-wise F1 loss curves during training.}
\label{losscurve}
\end{figure}

We compare the proposed VOILA with 3 SOTA methods on 7 different datasets, including single-organ and multi-organ segmentation. Table \ref{table1} lists the average Dice score of all labelled organs. The VOILA proposed in this paper is trained using contrastive loss, which enhances segmentation performance as the number of classes increases. With more classes and a higher number of negative samples for each class, the boundaries between categories in the representation space become more distinct, leading to more precise segmentation results. As shown in the table, VOILA performs better on Ts-v2, WORD, and AMOS datasets with a larger number of categories, whereas its performance on single-category datasets falls short of expectations, including Pancreas-CT and LiTS. Furthermore, since the experiments were constrained by a fixed number of epochs, methods with more parameters and more complex architectures exhibit redundancy during training, requiring longer to converge. In contrast, our method, with fewer parameters, converges more rapidly and still achieves competitive results. As illustrated in Figure \ref{losscurve}, VOILA’s F1 loss curve undergoes two notable stepwise decreases, reflecting its efficient convergence, and it achieves the fastest convergence among the evaluated methods.

\subsection{Evaluation without Fine-tuning}
The proposed VOILA method in this paper classifies voxels using cosine similarity rather than employing a fully connected layer for class mapping. Consequently, the optimization objective is focused on how the voxel encoder learns the physical structural features necessary to align the voxels with the text tokens in the representation space. Moreover, the inclusion of CAS module enhances the model's sensitivity to spatial information, allowing it to capture more generalized representations that are independent of specific datasets. To evaluate the model's generalizability, we used the parameters trained on the Ts-v2 dataset, which has the most categories, and tested the model on other datasets without any fine-tuning. Table \ref{table2} demonstrates that VOILA trained with contrastive loss on the Ts-v2 dataset, which includes a large number of classes, performs well across other datasets. Notably, it shows significant improvement in NSD and HD95 metrics, further validating the strong generalizability of the proposed method.

\subsection{Ablation Study}
To further validate the effectiveness of the proposed method, we conducted ablation experiments on the Ts-v2 dataset, as shown in Table \ref{table3}. The experiments focus on the Voxel-Language Interaction segmentation method and the impact of different sampling strategies. First, the standard segmentation method requires training a fully connected layer with a large number of parameters, demanding more data and iterations. In contrast, the Voxel-Language Interaction method leverages text tokens extracted by a pre-trained text encoder as classification benchmarks, resulting in faster convergence. Additionally, the fully connected layer is tailored to specific datasets, while text prompts are dataset-agnostic, allowing the model to learn more generalized representations, yielding better performance on other datasets without fine-tuning. In terms of sampling, the results of all sampling methods outperform the non-sampling one. Given the high number of background voxels diluting the influence of foreground voxels, sampling facilitates faster model convergence and helps address the class imbalance problem. However, using too few sampling voxels inevitably increases the number of iterations, making it crucial to select an appropriate sampling rate. Moreover, at the same sampling rate, the proposed CAS module outperforms random sampling because it can sense classification complexity, targeting difficult-to-segment regions more effectively. Together with the Vision-Language Interaction method, it enhances the model’s generalizability further.

\begin{table}[t!]
\centering
\caption{Average Dice scores for different sampling ratios on the Ts-v2 dataset.}
\label{table4}
\begin{tabular}{c|cccccc}
\textbf{Ratio} & 0.01 & 0.1 & 0.3 & 0.5 & 0.7 & N/A \\ \hline
\textbf{mDice} & 88.6 & 92.1 & 91.5 & 90.6 & 88.1 & 86.0 \\ 
\end{tabular}
\end{table}

\subsection{Visualisation}
Figure \ref{heatmaps} demonstrates the heatmaps generated by the CAS module during training. As training progresses, the CAS module increasingly focuses on specific regions of higher segmentation complexity, and then gradually refine and spread across finer details throughout the image. Figure \ref{results} displays examples of segmentation results, highlighting improved segmentation of edge regions achieved by the proposed method. The CAS module plays a critical role by focusing the model's attention on these areas of high segmentation complexity, such as organ boundaries, resulting in better performance in edge regions with the same number of iterations.

\section{Conclusion}
In this paper, we introduce a brand new universal CT segmentation methods called VOILA. We propose a Voxel-Language interaction segmentation method, enhancing the loss function and textual prompts to address class imbalance. Additionally, we design a Complexity-Aware Sampling module that dynamically selects more challenging voxels during training, promoting faster convergence and better segmentation results, particularly in edge regions. Experimental results demonstrate that our approach achieves competitive performance with fewer parameters and lower computational cost in 7 public datasets. Furthermore, the proposed method achieves significant improvements on datasets with a large number of classes and exhibits superior generalizability on other datasets without fine-tuning.

\section{Acknowledgments}
The authors sincerely thank Professor Xinmiao Sun for her invaluable suggestions during the conceptualization phase of this paper. The authors are also deeply grateful to Professor Chao Yao for his insightful and experienced advice during the revision process. This work was supported by the National Natural Science Foundation of China (No. 62273035) and the Interdisciplinary Research Project for Young Teachers of USTB (Fundamental Research Funds for the Central Universities, FRF-IDRY-22-025).

\bibliography{aaai25}

\end{document}